\documentclass[sigconf]{acmart}

\usepackage{bbold}  
\usepackage{pifont}
\usepackage{bm}
\usepackage{cleveref}
\usepackage{multirow}
\usepackage{colortbl} 
\usepackage{dsfont}

\AtBeginDocument{%
  }

\copyrightyear{2025}
\acmYear{2025}
\setcopyright{cc}
\setcctype{by}
\acmConference[MM '25]{Proceedings of the 33rd ACM International Conference on Multimedia}{October 27--31, 2025}{Dublin, Ireland}
\acmBooktitle{Proceedings of the 33rd ACM International Conference on Multimedia (MM '25), October 27--31, 2025, Dublin, Ireland}
\acmDOI{10.1145/3746027.3755339}
\acmISBN{979-8-4007-2035-2/2025/10}

\acmSubmissionID{mfp3301}

\begin{document}

\settopmatter{printacmref=false} 
\setcopyright{none}             
\renewcommand\footnotetextcopyrightpermission[1]{}  
\pagestyle{plain}   
\fancyhead{}        

\title{Slot Attention with Re-Initialization and Self-Distillation}

\author{Rongzhen Zhao}
\authornote{Corresponding Author.}
\orcid{0009-0000-3964-7336}
\affiliation{
\department{Department of Electrical Engineering and Automation}
\institution{Aalto University}
\city{Espoo}
\country{Finland}
}
\email{rongzhen.zhao@aalto.fi}

\author{Yi Zhao}
\orcid{0009-0002-9979-595X}
\affiliation{
\department{Department of Electrical Engineering and Automation}
\institution{Aalto University}
\city{Espoo}
\country{Finland}
}
\email{yi.zhao@aalto.fi}

\author{Juho Kannala}
\orcid{0000-0001-5088-4041}
\affiliation{
\department{Department of Computer Science}
\institution{Aalto University}
\city{Espoo}
\country{Finland}
}
\affiliation{
\department{Center for Machine Vision and Signal Analysis}
\institution{University of Oulu}
\city{Oulu}
\country{Finland}
}
\email{juho.kannala@aalto.fi}

\author{Joni Pajarinen}
\orcid{0000-0003-4469-8191}
\affiliation{
\department{Department of Electrical Engineering and Automation}
\institution{Aalto University}
\city{Espoo}
\country{Finland}
}
\email{joni.pajarinen@aalto.fi}

\renewcommand{\shortauthors}{Rongzhen Zhao, Yi Zhao, Juho Kannala, and Joni Pajarinen}

\begin{abstract}
Unlike popular solutions based on dense feature maps, Object-Centric Learning (OCL) represents visual scenes as sub-symbolic object-level feature vectors, termed slots, which are highly versatile for tasks involving visual modalities.
OCL typically aggregates object superpixels into slots by iteratively applying competitive cross attention, known as Slot Attention, with the slots as the query.
However, once initialized, these slots are reused naively, causing redundant slots to compete with informative ones for representing objects. This often results in objects being erroneously segmented into parts.
Additionally, mainstream methods derive supervision signals solely from decoding slots into the input's reconstruction, overlooking potential supervision based on internal information. 
To address these issues, we propose Slot Attention with re-Initialization and self-Distillation (DIAS): $\emph{i)}$ We reduce redundancy in the aggregated slots and re-initialize extra aggregation to update the remaining slots; $\emph{ii)}$ We drive the bad attention map at the first aggregation iteration to approximate the good at the last iteration to enable self-distillation.
Experiments demonstrate that DIAS achieves state-of-the-art on OCL tasks like object discovery and recognition, while also improving advanced visual prediction and reasoning. 
Our source code and model checkpoints are available on https://github.com/Genera1Z/DIAS.
\end{abstract}



\maketitle

\begin{figure}
\centering
\sffamily\scriptsize
\textbf{\textit{o1}}: Slots redundancy reduction yields limited boosts in object recognition, i.e., classification (left) and localization (right), while re-initialization boosts both object discovery and recognition.
\includegraphics[width=0.95\linewidth]{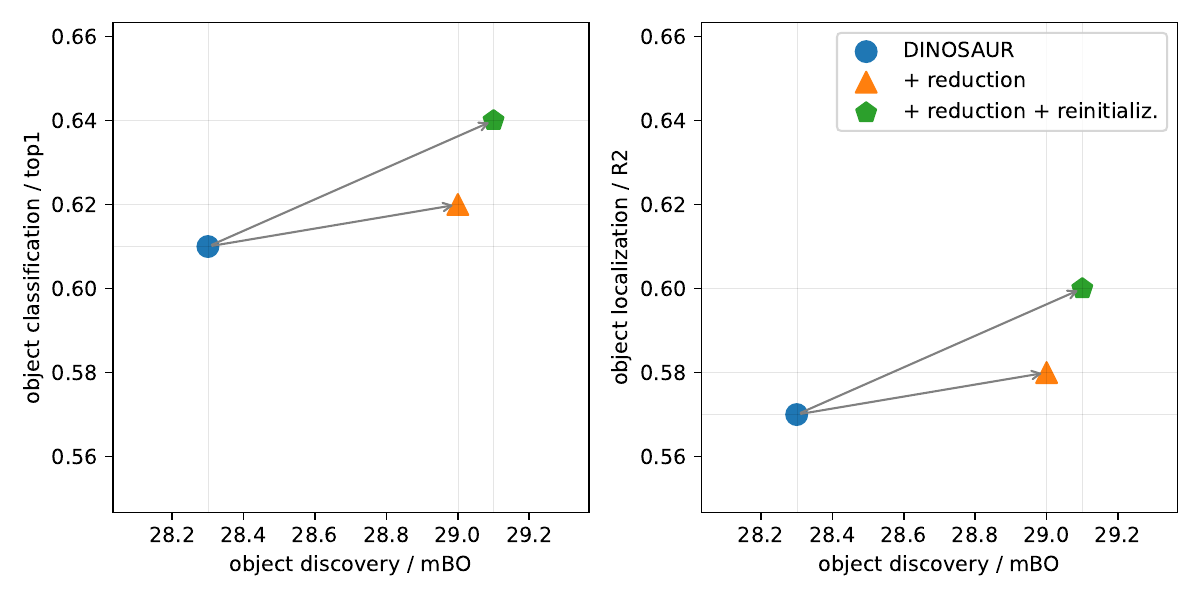}
\\\vspace{\baselineskip}
\textbf{\textit{o2}}: The decoding attention map does not always segment/discover objects better than the aggregation's (left); Attention maps of the 3rd aggregation iteration are almost always better than that of the first (right).
\includegraphics[width=0.95\linewidth]{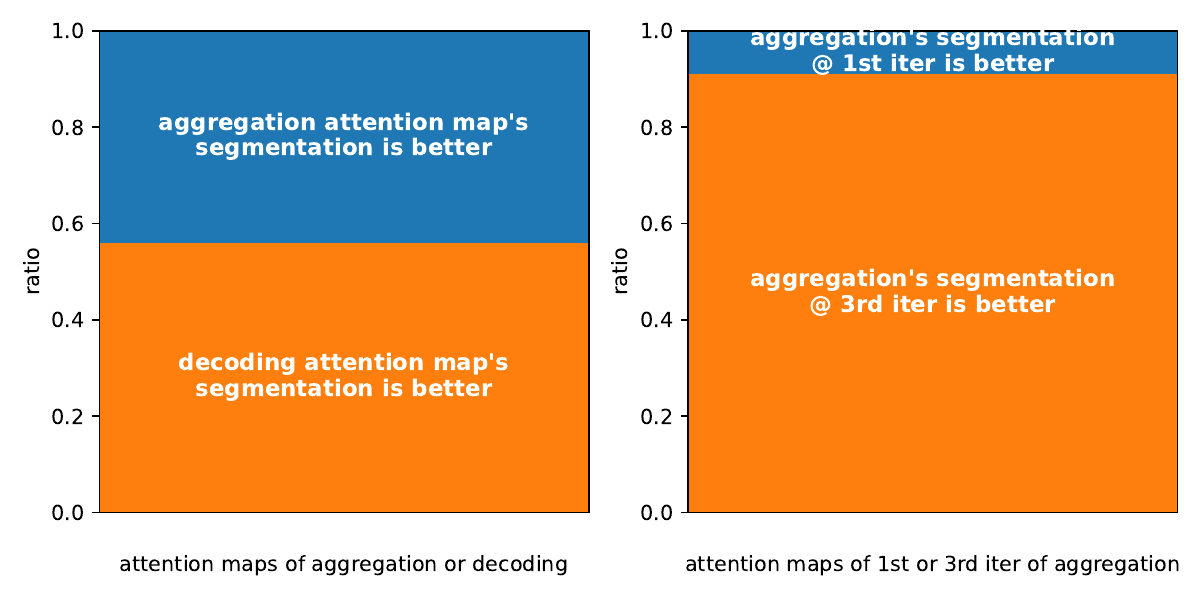}
\caption{\textmd{
Our method is inspired by the following two key observations.
(\textit{o1}) We re-initialize an extra aggregation to update the remaining slots after slots redundancy reduction, instead of decoding the remaining slots directly.
(\textit{o2}) We self-distill the attention map at the first iteration to approximate the almost-always-good attention at the last aggregation iteration, rather than from aggregation attention to decoding attention.
}}
\label{fig:teaser}
\end{figure}

\section{Introduction}
\label{sect:introduction}

Humans' vision system typically decomposes scenes into objects and allocates attention at the object level \cite{chen2012objectbasedattention, kahneman1992objectfiles}. This efficiently captures scene information, models object attributes and their interrelationships, and ultimately supports complex functionalities of prediction, reasoning, planning and decision-making \cite{bar2004visual}. Similarly, in Artificial Intelligence (AI), Object-Centric Learning (OCL) seeks to aggregate dense (super-)pixels of images or videos into sparse feature vectors, termed slots, representing corresponding objects in a sub-symbolic manner \cite{locatello2020slotattent}. In multimodal AI systems, object-centric representations can encode visual information more effectively and integrate with other modalities more flexibly than popular approaches based on dense feature maps \cite{zhang2024omgllava, wang2024omnidrive}.

Existing OCL typically follows the encoding-aggregation-decoding paradigm \cite{zhao2025vvo}. \textit{Firstly}, input image or video pixels are encoded into feature map; \textit{Secondly}, object super-pixels in the feature map are iteratively aggregated into corresponding feature vectors, i.e., slots, using competitive cross attention, i.e., Slot Attention \cite{locatello2020slotattent} or its variants \cite{biza2023isa, jia2023boqsa}; \textit{Lastly}, these slots are decoded to reconstruct the input in some format, providing the self-supervision signal.

However, (\textit{i1}) once initialized, these slots are iterated naively in the aggregation, and when there are fewer objects than the predefined number of slots, redundant slots would compete with informative ones for object representation \cite{aydemir2023solv, fan2024adaslot}, causing objects to be segmented into parts. 
Additionally, (\textit{i2}) mainstream methods \cite{singh2021slate, seitzer2023dinosaur, wu2023slotdiffuz} derive supervision signals solely from reconstructing the input using the aggregated slots, leaving internal information unutilized, e.g., the aggregation attention maps at latter iterations have better object segmentation, which can be utilized to provide extra supervision signal.

Issues (\textit{i1}) and (\textit{i2}) remain unresolved despite just a few ingenious attempts. 
(\textit{r1}) SOLV \cite{aydemir2023solv} and AdaSlot \cite{fan2024adaslot} remove possible redundancy in the aggregated slots, but without further updating the remaining.
As shown in \Cref{fig:teaser} \textit{o1}, the remaining slots, despite being informative, are already distracted by redundant slots and are still of low quality in object representation.
(\textit{r2}) SPOT \cite{kakogeorgiou2024spot} drives the aggregation attention map to approximate the whole decoding attention map, but without discarding the latter's bad compositions. As shown in \Cref{fig:teaser} \textit{o2}, the decoding attention map, despite being generally better, sometimes has lower accuracy in object segmentation than the aggregation attention map. This is why SPOT has to pretrain an extra copy of the model as the teacher and has to run both the teacher and student for their offline distillation, at least doubling the training cost.

Our solution is simple yet effective. 
(\textit{s1}) We use the remaining slots after slots redundancy reduction to re-initialize an extra aggregation. This improves slots' object representation quality, instead of directly decoding the remaining slots.
(\textit{s2}) We use the attention map at the last aggregation iteration to guide that at the first iteration, where the guidance is mostly always better. This enables self-distillation, requiring no extra model copy, no pretraining, and minimal additional cost.
Besides, (\textit{s3}) we implement a generalized auto-regressive (AR) decoder for OCL. Since existing AR decoders' fix-order flattening disrupts spatial correlations among superpixels \cite{singh2021slate, singh2022steve, kakogeorgiou2024spot}, we use random-order flattening to enforce the decoder to model spatial correlations.

In short, our contributions are 
(\textit{i}) re-initialized OCL aggregation,
(\textit{ii}) self-distillation within aggregation iterations,
(\textit{iii}) random auto-regressive decoding;
(\textit{iv}) new state-of-the-art performance.

\begin{figure*}
\centering
\includegraphics[width=\linewidth]{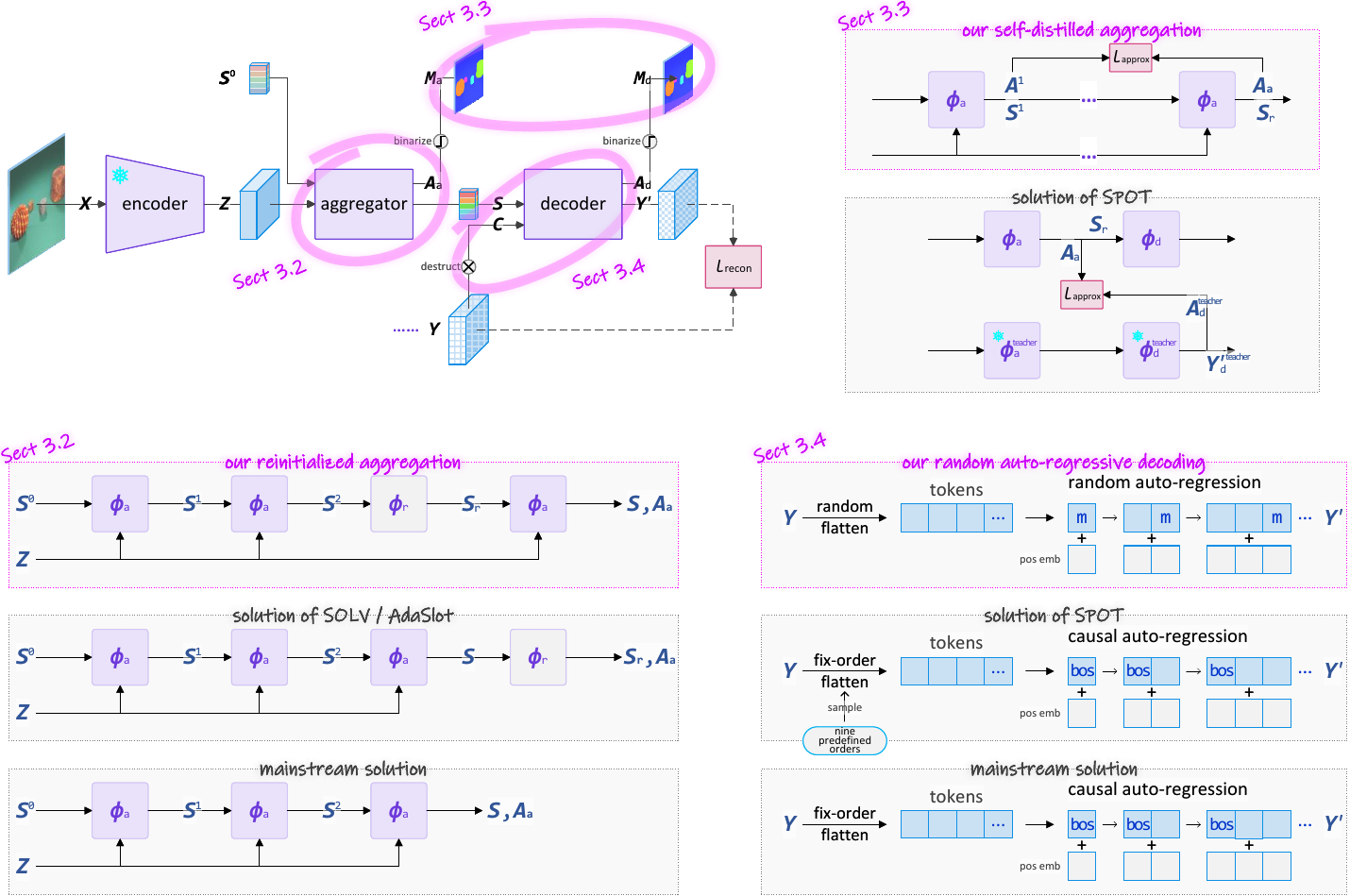}
\caption{\textmd{
Our DIAS introduces three novel designs: 
(\textit{i}) Reinitialized aggregation, which improves slots' object representation quality by re-initializing an extra aggregation to update the remaining slots after slots redundancy reduction; 
(\textit{ii}) Self-distilled aggregation, which obtains better internal supervision by approximating the almost-always-better attention map at the last aggregation iteration from that at the first aggregation iteration;
and (\textit{iii}) Random auto-regressive decoding, which enforces the decoder's modeling of spatial correlations by randomly flattening a 2-dimensional feature map into a 1-dimensional sequence.
On the top left is the common OCL architecture, which is adapted from \cite{zhao2025vvo}.
$\bm{\phi}_\mathrm{a}$ is OCL aggregator, $\bm{\phi}_\mathrm{d}$ is OCL decoder, and $\bm{\phi}_\mathrm{r}$ is slots redundancy reduction. \texttt{m} is the mask token, \texttt{bos} is the Begin-of-Sentence token, and \texttt{pos emb} stands for the position embedding tensor. All notations are defined in the marked sections.
}}
\label{fig:solution}
\end{figure*}

\section{Relate Work}
\label{sect:related_work}

\textbf{OCL aggregation}. 
The core module of OCL is Slot Attention \cite{locatello2020slotattent}, an iterative cross attention that takes some feature vectors as the competitive query, with the input's feature map as the key and value.
BO-QSA \cite{jia2023boqsa} improves the gradient flow on the query vectors for better optimization. 
ISA \cite{biza2023isa} realizes scaling, rotation and translation invariance in slots with inductive biases. 
While others learn Gaussian distributions to sample the query vectors, SAVi \cite{kipf2021savi} and SAVi++ \cite{elsayed2022savipp} project conditions like object bounding boxes into query vectors. 
These aggregators are all faced with the issue that redundant slots compete with informative slots to represent single objects as parts.
SOLV \cite{aydemir2023solv} merges similar slots after aggregation using agglomerative clustering by some threshold. AdaSlot \cite{fan2024adaslot} removes redundant slots after aggregation using a learned policy with some regularization.
These two recent methods only remove or merge slots but do not update the remaining ones. Our DIAS improves the aggregation by both reducing redundant slots and updating the remaining slots with extra re-initialized aggregation.

\textbf{OCL decoding}. 
There are mainly three types of decoding for OCL. 
Mixture-based decoders include CNN \cite{kipf2021savi, elsayed2022savipp}, MLP \cite{seitzer2023dinosaur} and SlotMixer \cite{zadaianchuk2024videosaur}. The former two broadcast each of the aggregated slots into some spatial dimension, decode them respectively and mix the final components as the reconstruction. This means that the computation is proportional to the number of slots \cite{seitzer2023dinosaur}, which is very expensive. The latter one mixes the slots with a Transformer decoder without self-attention on the query. Such a decoder has to work in lower slot channel dimensions \cite{zadaianchuk2024videosaur}, which limits object representation quality.
Auto-regression-based decoders include causal Transformer decoder \cite{singh2021slate, singh2022steve} and causal Transformer decoder with nine permutations \cite{kakogeorgiou2024spot}. They basically drive the slots to reconstruct the input's feature map with the causally masked input feature as the query, in a next-token prediction manner. However, flattening a 2-dimensional feature map into a 1-dimensional sequence in a fixed order loses the spatial relationship \cite{kakogeorgiou2024spot}.
Diffusion-based decoders include different conditional Diffusion models \cite{wu2023slotdiffuz, jiang2023lsd}. They basically recover the noise from the noise diffused input with slots as conditions, to drive slots to be informative. But, Diffusion models are very large and sensitive to train \cite{rombach2022latentdiffusion}.
Our DIAS realizes general auto-regression via arbitrary ordering.
In terms of the reconstruction target, there are also works that introduce quantization \cite{singh2021slate, singh2022steve, zhao2024gdr, zhao2024msf, zhao2025vvo}, but we focus on the non-quantized case.

\textbf{Knowledge distillation}. 
The classical offline knowledge distillation (KD) \cite{gou2021kd} requires a pretrained model as the teacher and a to-be-trained model as the student. The teacher is frozen and runs at the same time as the student to provide better representations as targets for the corresponding student representations.
There is also online distillation \cite{gou2021kd} where no pretraining of a teacher model is needed. A teacher model typically has the same architecture as the student and its weights are usually updated by exponentially moving averaging \cite{wiki2025ema} the student model.
Self-distillation \cite{gou2021kd} does not need an extra model as the teacher. Instead, intermediate representations in the model's higher layers are taken as the teacher of the lower layers. This requires the higher layers' representations to be better than the lower layers'.
KD has not been fully explored in the OCL setting and SPOT \cite{kakogeorgiou2024spot} is the only one we know for now. But SPOT's ``self-distillation`` is actually offline distillation, which consumes at least two times training cost.
Our DIAS realizes exact self-distillation by utilizing the definitively refined attention maps of the aggregation iterations, saving much cost.

\section{Proposed Method}
\label{sect:proposed_method}

We begin with the preliminary of object-centric learning, highlighting the issues in existing OCL methods. Then, we present our method, Slot Attention with re-Initialization and self-Distillation (DIAS). In \Cref{fig:solution}, our method is illustrated and compared with three recent studies that share similar ideas.

\subsection{Preliminary}
\label{sect:preliminary_and_issues}

Existing OCL models typically consist of three major modules: encoder, aggregator and decoder. We focus on the latter two.

Given the initial slots, i.e., query vectors $\bm{S}^0 \in \mathbb{R} ^ {s \times c}$, a Slot Attention module $\bm{\phi}_{\mathrm{a}}$ aggregates the feature map  $\bm{Z} \in \mathbb{R} ^ {h \times w \times c}$ of an input image or video frame, into slots $\bm{S} \in \mathbb{R} ^ {s \times c}$ after some iterations:
\begin{equation}
\label{eq:aggregation}
\bm{S}^{i}, \bm{A}^{i}_\mathrm{a} = \bm{\phi}_\mathrm{a} (\bm{S}^{i-1}, \bm{Z}) \quad i=1...i_\mathrm{a}
\end{equation}
\begin{equation}
\label{eq:aggregated_slots}
\bm{S} \equiv \bm{S} ^ {i_\mathrm{a}}
\end{equation}
where $i_\mathrm{a}$ is the total number of iterations for aggregation, and $i_\mathrm{a}=3$ for unconditional queries \cite{locatello2020slotattent} while $i_\mathrm{a}=1$ for conditional queries \cite{kipf2021savi}. $\bm{A}^{i}_\mathrm{a} \in \mathbb{R} ^ {s \times h \times w}$ is aggregation attention map and can be converted to object segmentation with $\mathrm{argmax(\cdot)}$ along dimension $s$. $\bm{\phi}_{\mathrm{a}}$ can be parameterized into different Slot Attention variants like in \citet{biza2023isa, jia2023boqsa}. 
We call $\bm{S}^i$ as ``query slots'' and $\bm{S}$ as ``aggregated slots''.
Note that $\bm{\phi}_\mathrm{a}$ enables all query vectors in $\bm{S}^i$ to compete with one another for some super-pixels in $\bm{Z}$.

To obtain the self-supervision signal, the aggregated slots $\bm{S}$ are decoded by a decoder $\bm{\phi}_\mathrm{d}$ to reconstruct the input $\bm{X}$ in some format $\bm{Y} \in \mathbb{R} ^ {h' \times w' \times c'}$, with some clue $\bm{C}$:
\begin{equation}
\label{eq:decoding}
\bm{Y}', \bm{A}_\mathrm{d} = \bm{\phi}_\mathrm{d} (\bm{C}, \bm{S})
\end{equation}
where $\bm{A}_\mathrm{d} \in \mathbb{R} ^ {s \times h' \times w'}$ is decoding attention map and can be converted to object segmentation masks by binarization; $\bm{Y}'$ is the reconstructed $\bm{Y}$; and $\bm{Y}$ can be $\bm{X}$ \cite{locatello2020slotattent}, $\bm{Z}$ \cite{seitzer2023dinosaur}, or $\bm{Z}$'s quantization \cite{zhao2025vvo}, etc. $\bm{C}$ is the destructed $\bm{Y}$, which varies across methods.

The supervision signal comes from the reconstruction loss, either Mean Squared Error (MSE) or Cross Entropy (CE):
\begin{equation}
\label{eq:mse}
l_\mathrm{recon} = \mathrm{MSE} (\bm{Y}', \bm{Y})
\end{equation}
\begin{equation}
\label{eq:ce}
l_\mathrm{recon} = \mathrm{CE} (\bm{Y}', \bm{Y})
\end{equation}
The choice depends on specific methods. Mixture- and diffusion-based decoding use \Cref{eq:mse}, while auto-regression-based decoding uses either \Cref{eq:mse} or \Cref{eq:ce}.

\begin{figure*}
\centering
\includegraphics[width=\linewidth]{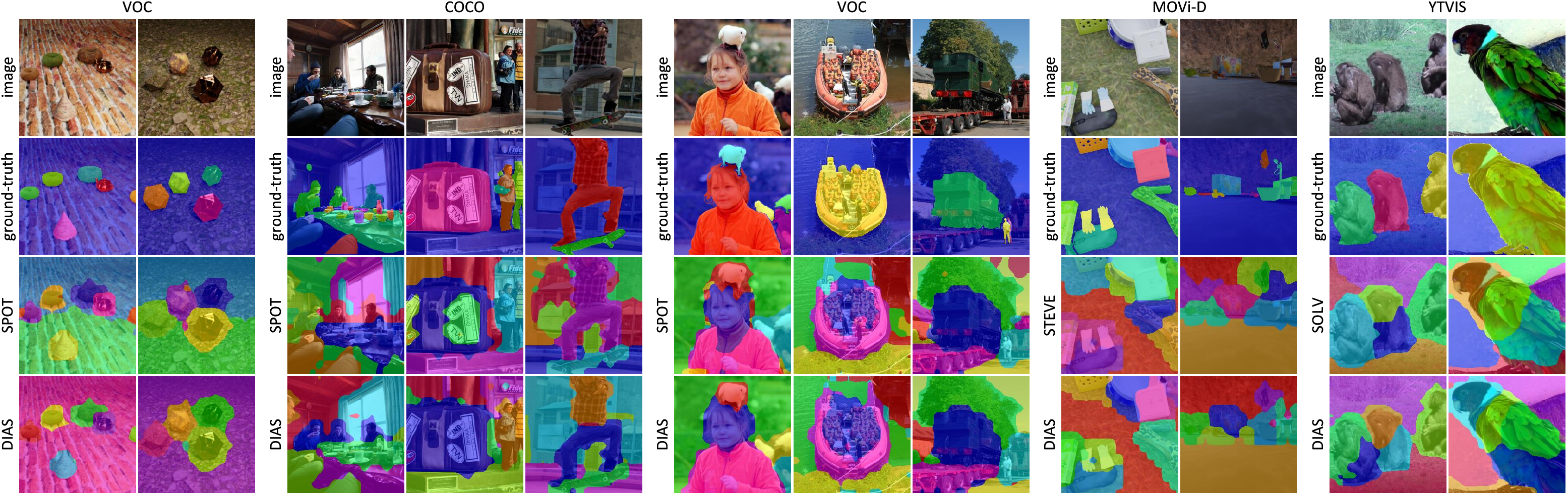}
\caption{\textmd{
Visualization of object discovery.
}}
\label{fig:qualitative}
\end{figure*}

\subsection{Re-Initialized Aggregation}
\label{sect:reinitialized_aggregation}

Here is the first issue we address. 

When there are fewer objects (plus the background) than the number of slots, redundant slots should be removed, or they will compete for object super-pixels and cause objects to be segmented into parts. SOLV \cite{aydemir2023solv} and AdaSlot \cite{fan2024adaslot} are the only two studies so far that explore this idea. They both conduct the redundancy reduction $\bm{\phi}_\mathrm{r}$ after the aggregation:
\begin{equation}
\label{eq:redundancy_removal}
\bm{S}_\mathrm{r}, \bm{M}_\mathrm{r} = \bm{\phi}_\mathrm{r} (\bm{S})
\end{equation}
\begin{equation}
\label{eq:masked_decoding}
\bm{Y}', \bm{A}_\mathrm{d} = \bm{\phi}_\mathrm{d} (\bm{C}, \bm{S}_\mathrm{r}, \bm{M}_\mathrm{r})
\end{equation}
where $\bm{S}_\mathrm{r}$ is in the same shape as $\bm{S}$ with redundant slots being zeroed out; $\bm{M}_\mathrm{r} \in \mathbb{R} ^ s$ tells the decoder $\bm{\phi}_\mathrm{d}$ which slots are zeroed. \Cref{eq:masked_decoding} is no different from \Cref{eq:decoding} except masking.

But as the slots aggregation is competitive, the remaining slots are already affected by those redundant slots and are still of low quality in object representing. 
As shown in \Cref{fig:teaser} \textit{o1}, with redundancy reduction solely, we yield some boosts on object discovery but very limited boosts on object recognition. 

Thus, we propose using the remaining slots after redundancy reduction to re-initialize an extra aggregation. Thus our aggregation and encoding are formulated as below:
\begin{equation}
\label{eq:our_aggregation1}
\bm{S}^{i}, \bm{A}^{i}_\mathrm{a} = \bm{\phi}_\mathrm{a} (\bm{S}^{i-1}, \bm{Z}) \quad i=1...i_\mathrm{a}-1
\end{equation}
\begin{equation}
\label{eq:our_redundancy_removal}
\bm{S}_{\mathrm{r}}, \bm{M}_\mathrm{r} = \bm{\phi}_\mathrm{r} (\bm{S}^{i_\mathrm{a}-1})
\end{equation}
\begin{equation}
\label{eq:our_aggregation2}
\bm{S}, \bm{A}_\mathrm{a} = \bm{\phi}_\mathrm{a} (\bm{S}_\mathrm{r}, \bm{Z}, \bm{M}_\mathrm{r})
\end{equation}
\begin{equation}
\label{eq:our_masked_decoding}
\bm{Y}', \bm{A}_\mathrm{d} = \bm{\phi}_\mathrm{d} (\bm{C}, \bm{S}, \bm{M}_\mathrm{r})
\end{equation}
Here \Cref{eq:our_aggregation1} is the initial aggregation, \Cref{eq:our_redundancy_removal} is the redundancy reduction, \Cref{eq:our_aggregation2} is the extra aggregation, and \Cref{eq:our_masked_decoding} is the masked decoding. 
For the initial aggregation, we use $i_\mathrm{a}-1$ iterations, and $i_\mathrm{a} \geq 2$.
For the redundancy reduction, we adopt SOLV's solution, agglomerative clustering \cite{sklearn2025agglomerat} by cosine distance given some threshold.
For the extra aggregation, we apply the redundancy mask $\bm{M}_\mathrm{r}$ to the query-key attention logits so as to discard the effect of those zeroed-out slots \cite{fan2024adaslot}.
For the masked decoding, we follow the solutions from SOLV and AdaSlot.

As shown in \Cref{fig:teaser} \textit{o1}, with further updating the remaining slots after redundancy reduction by re-initializing extra aggregation, we can yield boosts in both object discovery and recognition.

\begin{table*}[]
\centering
\setlength{\aboverulesep}{0pt}  
\setlength{\belowrulesep}{0pt}  
\newcommand{\tss}[1]{\scalebox{0.8}{\texttt{#1}}}
\newcommand{\cg}[1]{\textcolor{green}{#1}}
\newcommand{\std}[1]{\scalebox{0.4}{±#1}}

\begin{tabular}{ccccccccccccc}
\hline
& \multicolumn{4}{c}{ClevrTex {\tiny \#slot=11}} & \multicolumn{4}{c}{COCO {\tiny \#slot=7}} & \multicolumn{4}{c}{VOC {\tiny \#slot=6}} \\
\arrayrulecolor{gray}
\cmidrule(lr){2-5} \cmidrule(lr){6-9} \cmidrule(lr){10-13}
\arrayrulecolor{black}
& ARI & ARI\textsubscript{fg} & mBO & mIoU & ARI & ARI\textsubscript{fg} & mBO & mIoU & ARI & ARI\textsubscript{fg} & mBO & mIoU \\
\cmidrule(){1-13}
SLATE      & 17.4\std{2.9}  & 87.4\std{1.7} & 44.5\std{2.2} & 43.3\std{2.4} & 17.5\std{0.6} & 28.8\std{0.3} & 26.8\std{0.3} & 25.4\std{0.3} & 18.6\std{0.1} & 26.2\std{0.8} & 37.2\std{0.5} & 36.1\std{0.4} \\
DINOSAUR   & 50.7\std{24.1} & 89.4\std{0.3} & 53.3\std{5.0} & 52.8\std{5.2} & 18.2\std{1.0} & 37.0\std{1.2} & 28.3\std{0.5} & 26.9\std{0.5} & 21.5\std{0.7} & 36.2\std{1.3} & 40.6\std{0.6} & 39.7\std{0.6} \\
SlotDiffusion & 66.1\std{1.3}  & 82.7\std{1.6} & 54.3\std{0.5} & 53.4\std{0.8} & 17.7\std{0.5} & 29.0\std{0.1} & 27.0\std{0.4} & 25.6\std{0.4} & 17.0\std{1.2} & 21.7\std{1.8} & 35.2\std{0.9} & 34.0\std{1.0} \\
\arrayrulecolor{gray}
\cmidrule(lr){1-1} \cmidrule(lr){2-5} \cmidrule(lr){6-9} \cmidrule(lr){10-13}
AdaSlot     & -- & -- & -- & -- & 19.1\std{2.3} & 35.2\std{2.0} & 29.4\std{1.2} & 27.5\std{0.8} & -- & -- & -- & -- \\
SPOT        & 25.6\std{1.4} & 77.1\std{0.7} & 48.3\std{0.5} & 46.4\std{0.6} & 20.0\std{0.5} & 40.0\std{0.7} & 30.2\std{0.3} & 28.6\std{0.3} & 20.3\std{0.7} & 33.5\std{1.1} & 40.1\std{0.5} & 38.7\std{0.7} \\
\cmidrule(lr){1-1} \cmidrule(lr){2-5} \cmidrule(lr){6-9} \cmidrule(lr){10-13}
DIAS\tss{image} & \cg{81.6}\std{1.2} & \cg{80.5}\std{0.7} & \cg{64.2}\std{0.1} & \cg{62.4}\std{0.2} & \cg{22.1}\std{0.8} & \cg{42.7}\std{0.2} & \cg{32.8}\std{0.1} & \cg{30.1}\std{0.3} & \cg{27.3}\std{4.6} & \cg{34.6}\std{2.5} & \cg{44.8}\std{9.8} & \cg{42.8}\std{1.7} \\
\arrayrulecolor{black}
\hline
\end{tabular}

\caption{\textmd{
Object discovery on images. 
Using DINO2 ViT s/14 for OCL encoding; input resolution is 256$\times$256 (224$\times$224).
}}
\label{tab:objectdiscovery-images}
\end{table*}

\subsection{Self-Distilled Aggregation}
\label{sect:selfdistilled_aggregation}

Here is the second issue we address. 

The key to realizing self-distillation or simply distillation is that the teacher should be better than the student during training (of course after training the student can be better than the teacher).
SPOT \cite{kakogeorgiou2024spot} resorts to pretrain an extra copy of the model as the teacher to realize such performance gap.

However, the decoding attention map $\bm{A}_{\mathrm{a}}$ generally exhibits better object segmentation than the aggregation attention map $\bm{A}_{\mathrm{d}}$, but not always, as shown in \Cref{fig:teaser} \textit{o2} left. 
In contrast, the attention map at the last aggregation iteration is almost always better than that at the first iteration.
Thus, we can enforce additional supervision by forcing the first aggregation attention map to approximate the attention map at the last aggregation iteration.

SPOT \cite{kakogeorgiou2024spot} utilizes all the decoder attention map as the target without selection. And if some decoding attention compositions are worse than the corresponding aggregation attention, then such approximation would harm the performance. 
This is also why they have to pretrain an extra copy of the model and use it as the teacher to guide the training of the (student) model. Such offline distillation at least doubles the training cost. 
\begin{equation}
\label{eq:spot_distill_match}
\bm{M}_\mathrm{a}^* = \mathrm{Hungarian}( \mathrm{bin}_s (\bm{A}_\mathrm{a}), \mathrm{bin}_s (\bm{A}_\mathrm{d}^\mathrm{teacher}) )
\end{equation}
\begin{equation}
\label{eq:spot_distill_approx}
l_\mathrm{approx} = \mathrm{CE} ( \bm{A}_\mathrm{a}, \bm{M}_\mathrm{a}^* )
\end{equation}
where $\mathrm{bin}_s$ is binarization along dimension $s$, with maximum value as one and others as zeros; $\mathrm{Hungarian}(\cdot)$ is Hungarian matching; $l_\mathrm{approx}$ is the approximation loss.


To address this issue, we gain the signal for self-distillation from the intrinsic performance gap between the last aggregation iteration and the first.
As shown in \Cref{fig:teaser} \textit{o2} right, in 90\% cases, the attention map at the last aggregation iteration is better than that at the first iteration.

And our solution is quite simple:
\begin{equation}
\label{eq:dias_distill_match}
\bm{M}_\mathrm{a}^* = \mathrm{Hungarian}( \mathrm{bin}_s (\bm{A}_\mathrm{a}^1), \mathrm{bin}_s (\bm{A}_\mathrm{a}) )
\end{equation}
\begin{equation}
\label{eq:dias_distill_approx}
l_\mathrm{approx} = \mathrm{CE} ( \bm{A}_\mathrm{a}^1, \bm{M}_\mathrm{a}^* )
\end{equation}
where $\mathrm{bin}_s$ is binarization along dimension $s$, with maximum value as one and others as zeros; $\mathrm{Hungarian}(\cdot)$ is Hungarian matching; $l_\mathrm{approx}$ is the approximation loss; $\bm{A}_\mathrm{a}^1$ is the attention map at the first aggregation iteration while $\bm{A}_\mathrm{a}$ is the attention map at the last aggregation -- after re-initialization defined in \Cref{sect:reinitialized_aggregation}.

As the teacher $\bm{A}_\mathrm{a}$ is better than $\bm{A}_\mathrm{a}^1$ in most cases, which is a intrinsic performance gap, we do not need to pre-train a model as the teacher to produce such definitive performance gap.

\subsection{Random Auto-Regressive Decoding}
\label{sect:randomized_decoding}

We also improve the decoding.

As discussed in \Cref{sect:related_work} ``OCL decoding'', three types OCL decoding all have their disadvantages. 
Mixture-based decoders like CNN \cite{kipf2021savi, elsayed2022savipp} and MLP \cite{seitzer2023dinosaur} are too computationally expensive, while SlotMixer \cite{zadaianchuk2024videosaur} is efficient but has limited object representation quality. 
Diffusion-based decoders \cite{jiang2023lsd, wu2023slotdiffuz} are expensive and sensitive to train.
Auto-regression-based decoders like Transformer decoder \cite{singh2021slate, singh2022steve} are very efficient, but their next-token prediction has to flatten the 2-dimensional super-pixels into 1-dimensional in a fixed order, thus they do not utilize the spatial correlation among super-pixels.
Transformer9 \cite{kakogeorgiou2024spot} utilizes nine handcrafted permutations $\{ \bm{I}^j | j=1...9 \}$ to better capture spatial dependencies:
\begin{equation}
\label{eq:tfd9_bos}
\bm{Y}_\mathrm{bos}^j = \mathrm{concat}_n (\bm{E}_\mathrm{bos}^j, \mathrm{gather}_n (\mathrm{flatten}_{h,w} (\bm{Y}), \bm{I}^j )[:-1] )
\end{equation}
\begin{equation}
\label{eq:tfd9_dec}
{\bm{Y}^j}' = \bm{\phi}_\mathrm{d} ( \bm{Y}^j_\mathrm{bos}, \bm{M}_\mathrm{causal} )
\end{equation}
where $\bm{I}^j \in \mathbb{R} ^ n$ is the $j$th handicraft super-pixel ordering, with $n = h \times w$; $\bm{E}^j_\mathrm{bos} \in \mathbb{R}^c$ is the $j$th Beginning-of-Sentence (BoS) token \cite{kakogeorgiou2024spot}; $\mathrm{flatten}_{h,w} (\cdot)$ flattens height and width; $\mathrm{gather}_n (\cdot, \cdot)$ select the tensor along dimension $n$ given index tensor $\bm{I}^j$; $\mathrm{concat}_n (\cdot, \cdot)$ concatenates two tensors along dimension $n$; $\bm{Y}^j_\mathrm{bos} \in \mathbb{R} ^ {n \times c}$ is the sequence prepended with BoS; $\bm{M}_\mathrm{causal} \in \mathbb{R} ^ {n \times n}$ is the typical causal mask \cite{kakogeorgiou2024spot}; and $\bm{\phi}_\mathrm{d}$ is parameterized as a Transformer decoder.

The spatial relationships among super-pixels cannot be sufficiently captured by these nine handcrafted orderings. Therefore, we propose a generalized auto-regressive decoding scheme that accommodates arbitrary super-pixel orderings.
\begin{equation}
\label{eq:our_rand_len}
\quad n^\# \sim \mathcal{U}\{0, 1, \dots, n\}, ~~n = h \times w,
\end{equation}
\begin{equation}
\label{eq:our_shuffle_idx}
\bm{I} = \mathrm{shuffle}(\bm{I}^0)
\end{equation}
\begin{equation}
\label{eq:our_shuffle_pe}
\bm{E}_\mathrm{p}^\# = \mathrm{gather}_n (\mathrm{flatten}_{h,w}(\bm{E}_\mathrm{p}), \bm{I} [:n^\#+1])
\end{equation}
\begin{equation}
\label{eq:our_shuffle_y}
\bm{Y}^\# = \mathrm{concat}_{n^\#} ( \mathrm{gather}_n(\mathrm{flatten}_{h,w}(\bm{Y}), \bm{I} [:n^\#]), \bm{E}_\mathrm{m})
\end{equation}
\begin{equation}
\label{eq:our_rand_ar_dec}
{\bm{Y}^\#}' = \bm{\phi}_\mathrm{d}( \bm{E}^\#_\mathrm{p} + \bm{Y}^\#)
\end{equation}
\Cref{eq:our_rand_len} obtains a random sequence length $n^\#$; \Cref{eq:our_shuffle_idx} samples an arbitrary sequence order; \Cref{eq:our_shuffle_pe} shuffles the position embedding tensor $\bm{E}_\mathrm{p} \in \mathbb{R} ^ {h \times w \times c}$, while \Cref{eq:our_shuffle_y} shuffles the target tensor $\bm{Y}$, with mask token $\bm{E}_\mathrm{m} \in \mathbb{R}^c$ indicating the next token to predict; and \Cref{eq:our_rand_ar_dec} takes the positional information and the known tokens to predict the next token.

Accordingly, the reconstruction loss is modified as below:
\begin{equation}
\label{eq:our_recon_loss}
l_\mathrm{recon} = \mathrm{MSE}( {\bm{Y}^\#}', \bm{Y}^\# )
\end{equation}

Our random auto-regression compels the decoder $\bm{\phi}_\mathrm{d}$ to reconstruct the target $\bm{Y}$ using whatever subset of known super-pixels from arbitrary positions in the feature map. This, combined with the former two techniques, further boosts OCL performance.

\section{Experiment}
\label{sect:experiment}

We conduct the following experiments using three random seeds.
We evaluate object representation quality through object discovery and recognition, as well as downstream tasks, including video prediction and reasoning.

\subsection{Object Discovery}
\label{sect:object_discovery}

The attention maps of OCL aggregation and decoding are the segmentation of the objects and background that are discovered by slots. We measure these segmentation using typical metrics, i.e., ARI (Adjusted Rand Index) \cite{sklearn2025ari}, ARI\textsubscript{fg} (foreground Adjusted Rand Index), mBO (mean Best Overlap) \cite{uijlings2013selectivesearch} and mIoU (mean Intersection-over-Union) \cite{sklearn2025miou}. ARI measures more about larger areas, i.e., the background; mBO only measures the best overlapped segmentation; mIoU is the most strict measurement.

For baselines, we use the models from \cite{zhao2025vvo}, which reproduce very strong baselines by unifying representative OCL models all with current best practices. Among them, SLATE, STEVE, DINOSAUR, SlotDiffusion and SPOT are designed for images; STEVE and VideoSAUR are for videos; VVO with different decoders support images or videos. We also include AdaSlot (for images) \cite{fan2024adaslot} and SOLV (for videos) \cite{aydemir2023solv}.

For datasets, we use ClevrTex (the main set for training and the out-of-distribution set for evaluation) \cite{karazija2clevrtex}, COCO (instance segmentation) \cite{lin2014coco}, VOC \cite{everingham2010voc}, MOVi-D \cite{google2025movi} and YTVIS (high-quality) \cite{syscv2022hqytvis}. Among them, ClevrTex is synthetic images with complex textures; COCO and VOC are real-world images; MOVi-D is a set of challenging synthetic videos with up to 20 objects per scene; YTVIS is real-world YouTube videos.

As shown in \Cref{tab:objectdiscovery-images}, with unconditional query, our DIAS outperforms all the baselines on both synthetic and real-world images in object discovery under all the metrics. In fact, DIAS achieves new state-of-the-art performance.
As shown in \Cref{tab:objectdiscovery-videos}, with conditional query, DIAS the temporal version surpasses STEVE by a large margin on synthetic videos; With unconditional query, DIAS defeats the state-of-the-art video OCL model.
As shown in \Cref{tab:objectdiscovery-images-generalimprover}, combined with the general improver VVO \cite{zhao2025vvo}, DIAS pushes state-of-the-art performance forward even further.

Specifically, only SOLV, AdaSlot and DIAS, as well as DIAS temporal, have the redundant slot reduction operation, but DIAS is significantly better than the others. 
Besides, only SPOT and DIAS have the distillation operation, but SPOT has to pretrain an extra copy of the model as the teacher and then run both the teacher and a new copy of model simultaneously to do the distillation. By contrast, DIAS only needs to run one single copy of model without any pretraining, and thus, as shown in \Cref{tab:objectdiscovery_trainingcost}, only consumes average-level training resources. 

\begin{table}[]
\centering
\setlength{\aboverulesep}{0pt}  
\setlength{\belowrulesep}{0pt}  
\newcommand{\tss}[1]{\scalebox{0.8}{\texttt{#1}}}
\newcommand{\cg}[1]{\textcolor{green}{#1}}
\newcommand{\std}[1]{\scalebox{0.4}{±#1}}

\begin{tabular}{ccccccccccccc}
\hline
& ARI & ARI\textsubscript{fg} & mBO & mIoU \\
\arrayrulecolor{gray}
\cmidrule(lr){2-5}
\arrayrulecolor{black}
& \multicolumn{4}{c}{MOVi-D {\tiny \#slot=21 conditional}} \\
\cmidrule(){1-5}
STEVE   & 31.2\std{3.8} & 62.9\std{5.1} & 22.2\std{2.2} & 19.9\std{2.4} \\
DIAS\tss{video}    & \cg{37.2}\std{3.5} & \cg{64.7}\std{3.7} & \cg{25.9}\std{2.4} & \cg{22.7}\std{2.6} \\
\hline
& \multicolumn{4}{c}{YTVIS {\tiny \#slot=7 \#step=20}} \\
\cmidrule(){1-5}
VideoSAUR   & 34.6\std{0.5} & 48.6\std{0.7} & 31.4\std{0.3} & 31.2\std{0.3} \\
SOLV    & 36.2\std{1.1} & 50.4\std{1.2} & 30.7\std{0.8} & 31.0\std{0.7} \\
DIAS\tss{video} & \cg{38.7}\std{1.0} & \cg{52.1}\std{0.4} & \cg{33.3}\std{0.7} & \cg{34.6}\std{0.6} \\
\arrayrulecolor{black}
\hline
\end{tabular}

\caption{\textmd{
Object discovery on videos. Using DINO2 ViT s/14 for OCL encoding; input resolution is 256$\times$256 (224$\times$224).
}}
\label{tab:objectdiscovery-videos}
\end{table}

\begin{table}[]
\centering
\setlength{\aboverulesep}{0pt}  
\setlength{\belowrulesep}{0pt}  
\newcommand{\tss}[1]{\scalebox{0.8}{\texttt{#1}}}
\newcommand{\cg}[1]{\textcolor{green}{#1}}
\newcommand{\std}[1]{\scalebox{0.4}{±#1}}

\begin{tabular}{ccccccccccccc}
\hline
& \multicolumn{4}{c}{COCO {\tiny \#slot=7}} \\
\arrayrulecolor{gray}
\cmidrule(lr){2-5}
\arrayrulecolor{black}
& ARI & ARI\textsubscript{fg} & mBO & mIoU \\
\cmidrule(){1-5}
VVO\tss{Tfd}    & 21.1\std{2.1} & 31.5\std{1.1} & 29.6\std{0.7} & 28.2\std{0.8} \\
VVO\tss{Mlp}    & 19.2\std{0.4} & 35.9\std{0.6} & 28.7\std{0.3} & 27.4\std{0.3} \\
VVO\tss{Dfz}    & 18.3\std{0.4} & 28.7\std{1.0} & 27.2\std{0.1}  & 25.8\std{0.1}\\
\arrayrulecolor{gray}
\cmidrule(lr){1-1} \cmidrule(lr){2-5}
DIAS\tss{image}+VVO & \cg{22.9}\std{1.7} & \cg{43.1}\std{0.8} & \cg{33.2}\std{0.7} & \cg{31.7}\std{0.5} \\
\arrayrulecolor{black}
\hline
\end{tabular}

\caption{\textmd{
Combined with technique VVO, DIAS pushes SotA forward even further. Using DINO2 ViT s/14 for OCL encoding; input resolution is 256$\times$256 (224$\times$224).
}}
\label{tab:objectdiscovery-images-generalimprover}
\end{table}

\begin{table}[]
\centering
\setlength{\aboverulesep}{0pt}  
\setlength{\belowrulesep}{0pt}  
\newcommand{\tss}[1]{\scalebox{0.8}{\texttt{#1}}}
\newcommand{\cg}[1]{\textcolor{green}{#1}}
\newcommand{\std}[1]{\scalebox{0.4}{±#1}}

\begin{tabular}{ccccc}
\hline
& \multicolumn{4}{c}{COCO {\tiny \#slot=7}} \\
\arrayrulecolor{gray}
\cmidrule(lr){2-5}
& \multicolumn{2}{c}{pretraining} & \multicolumn{2}{c}{training} \\
\cmidrule(lr){2-3} \cmidrule(lr){4-5}
& memory & time & memory & time \\
\arrayrulecolor{black}
\cmidrule(){1-5}
SPOT            & 6.4 GB & 11.6 h & 7.0 GB & 12.1 h \\
DIAS\tss{image} & \cg{N/A}    & \cg{N/A}  & \cg{6.5 GB} & \cg{12.0 h} \\
\arrayrulecolor{black}
\hline
\end{tabular}

\caption{\textmd{
DIAS spends merely no more than 1/2 training cost compared with SPOT. Using one RTX 3080; auto-mixed precision.
}}
\label{tab:objectdiscovery_trainingcost}
\end{table}

\subsection{Object Recognition}
\label{sect:object_recognition}

With slots representing different objects or background, we can further evaluate how much object information they aggregate. In other studies, this is also called set prediction \cite{locatello2020slotattent} or object property prediction \cite{fan2024adaslot}. Following the routine of \cite{seitzer2023dinosaur}, we reuse the trained models from \Cref{sect:object_discovery} to extract the real-world image dataset COCO into slots representation, and train a small MLP to take each slot as input to predict the corresponding object's class label and bounding box. We use metrics top1 \cite{sklearn2025topkacc} and R2 \cite{sklearn2025r2} to measure the classification and regression performance.

As shown in \Cref{tab:objectrecognition-coco}, whether using the general improve VVO, our method achieves obviously better classification accuracy and bounding box regression accuracy on those real-world images. Also considering its object discovery superiority, we can assert that DIAS can represent visual scenes with better object representation quality than all the baselines.

\begin{table}[]
\centering
\setlength{\aboverulesep}{0pt}  
\setlength{\belowrulesep}{0pt}  
\newcommand{\tss}[1]{\scalebox{0.8}{\texttt{#1}}}
\newcommand{\cg}[1]{\textcolor{green}{#1}}
\newcommand{\std}[1]{\scalebox{0.4}{±#1}}

\begin{tabular}{ccccc}
\hline
&&& \multicolumn{2}{c}{COCO {\tiny \#slot=7}} \\
\arrayrulecolor{gray}
\cmidrule(lr){4-5}
&&& class top1$\uparrow$ & bbox R2$\uparrow$ \\
\arrayrulecolor{black}
\cmidrule(){1-5}
DINOSAUR & + & MLP          & 0.61\std{0.0} & 0.57\std{0.1} \\
SPOT & + & MLP              & 0.67\std{0.0} & 0.62\std{0.1} \\
DIAS\tss{image} & + & MLP   & \cg{0.70}\std{0.0} & \cg{0.63}\std{0.0} \\
\arrayrulecolor{black}
\hline
\end{tabular}

\caption{\textmd{
Object recognition on COCO.
}}
\label{tab:objectrecognition-coco}
\end{table}

\subsection{Visual Prediction and Reasoning}
\label{sect:visual_prediction_and_reasoning}

\begin{figure}
\centering
\includegraphics[width=\linewidth]{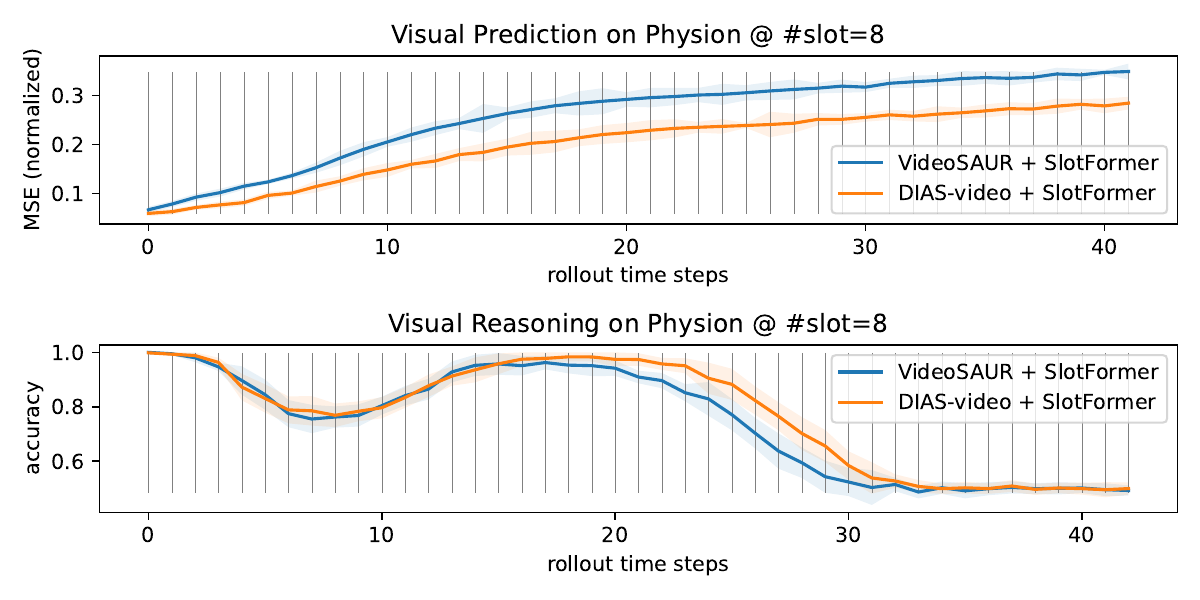}
\caption{\textmd{
Visual prediction (upper) and reasoning (lower).
}}
\label{fig:objectpredictreason}
\end{figure}

With improved object representation, our DIAS also helps the downstream tasks, like visual prediction and reasoning. 
Following the rountine from \cite{wu2022slotformer}, we firstly train our DIAS and other baselines on Physion \cite{bear2021physion}, which is a synthetic video dynamics modeling dataset, and represent the dataset as slots. Then we train an object-centric world model SlotFormer \cite{wu2022slotformer} to auto-regressively predict future frames of slots, given initial five frames of slots as the input. And based on such predictions, we further train a reasoning model to classify whether the stimuli objects will contact finally.

As shown in \Cref{fig:objectpredictreason} upper, in terms of visual prediction, measured by L2 norm normalized MSE, our combination DIAS+SlotFormer accumulates errors significantly slower than the baseline combination VideoSAUR+SlotFormer.
As shown in \Cref{fig:objectpredictreason} lower, in terms of visual reasoning, measured by binary classification accuracy, our combination DIAS+SlotFormer achieves better overall reasoning accuracy thanks to the much lower prediction error.

By the way, the reasoning accuracy goes down-up-down because: Early on, the model's predictions are accurate, allowing the classifier to quickly infer the final state and achieve high accuracy. As prediction errors start to accumulate, these errors distract the classifier and reduce accuracy. When more frames are predicted, the model receives additional (though noisy) evidence near the end, which boosts accuracy temporarily--until the accumulated error becomes too great and overwhelms the useful information.

\subsection{Ablation Study}

\begin{table}[]
\centering
\newcommand{\std}[1]{\scalebox{0.4}{±#1}}
\setlength{\aboverulesep}{0pt}  
\setlength{\belowrulesep}{0pt}  

\begin{tabular}{cccccc}
\hline
redundancy  & re-       & self-     & rand ar   & \multicolumn{2}{c}{COCO \#slot=7} \\
\arrayrulecolor{gray}
\cmidrule(lr){5-6}
\arrayrulecolor{black}
reduction   & initializ.& distill.  & decoding  & ARI       & ARI\textsubscript{fg} \\
\cmidrule(lr){1-4} \cmidrule(){5-6}
\checkmark  & \checkmark & \checkmark & \checkmark & 22.1\std{0.8} & 42.7\std{0.2}  \\
\arrayrulecolor{gray}
\cmidrule(lr){1-4} \cmidrule(lr){5-6}
\checkmark  &           & \checkmark & \checkmark & 21.3\std{0.8} & 42.0\std{0.4}   \\
\cmidrule(lr){1-4} \cmidrule(lr){5-6}
&       & \checkmark & \checkmark & 20.6\std{0.7} & 41.5\std{0.3} \\
\cmidrule(lr){1-4} \cmidrule(lr){5-6}
\checkmark & \checkmark &           & \checkmark & 21.4\std{1.7} & 41.8\std{0.5} \\
\cmidrule(lr){1-4} \cmidrule(lr){5-6}
\checkmark & \checkmark & \checkmark &          & 19.9\std{1.1} & 42.1\std{0.7} \\
\arrayrulecolor{black}
\hline
\end{tabular}

\caption{\textmd{
Effects of four techniques.
}}
\label{tab:ablat_overall}
\end{table}

\begin{table}[]
\centering
\newcommand{\std}[1]{\scalebox{0.4}{±#1}}
\setlength{\aboverulesep}{0pt}  
\setlength{\belowrulesep}{0pt}  

\begin{tabular}{cccc}
\hline
&           & \multicolumn{2}{c}{COCO \#slot=7} \\
\arrayrulecolor{gray}
\cmidrule(lr){3-4}
\arrayrulecolor{black}
redundancy reduction    & re-initializ. & class top1    & bbox R\textsuperscript{2} \\
\cmidrule(lr){1-4}
\checkmark & \checkmark & 0.70\std{0.0}  & 0.63\std{0.0} \\
\arrayrulecolor{gray}
\cmidrule(lr){1-2} \cmidrule(lr){3-4}
\checkmark &            & 0.68\std{0.0}  & 0.62\std{0.0} \\
\cmidrule(lr){1-4}
\multicolumn{2}{c}{\textcolor{gray}{SPOT as a reference}} & \textcolor{gray}{0.67\std{0.0}}  & \textcolor{gray}{0.62\std{0.1}} \\
\arrayrulecolor{black}
\hline
\end{tabular}

\caption{\textmd{
Effects of re-initialization.
}}
\label{tab:ablat_reinitializ}
\end{table}


Here we ablate our model to evaluate how much our three techniques contribute to the performance boosts respectively. We conduct the ablation on COCO using both object discovery and recognition tasks.

As shown in \Cref{tab:ablat_overall} and \Cref{tab:ablat_reinitializ}, redundancy reduction in aggregated slots, which is not our contribution, is important to the performance. But, this does not mean that our technique re-initialization is not necessary; instead, re-initialization builds on the reduction and improves it even further, especially in object recognition. 


As shown in \Cref{tab:ablat_overall}, our arbitrary-order auto-regressive decoding surely contributes positively to our DIAS model architecture.
Specifically, our self-distillation is also beneficial to both the background segmentation performance, i.e., ARI, and the foreground segmentation performance, i.e., ARI\textsubscript{fg}.
Our random AR decoding contributes much more to the background segmentation.

\section{Conclusion}
\label{sect:conclusion}

We explored three techniques in OCL aggregation and decoding modules, i.e., re-initialization in aggregation, self-distillation from the last aggregation attention to the first aggregation attention, and arbitrary-order auto-regressive decoding. With these techniques, our DIAS achieves new state-of-the-art in OCL tasks on both images and videos. The basis of our re-initialization still relies on manual threshold and so does our self-distillation. These could be future works to improve DIAS further.

\begin{acks}
We acknowledge the support of Finnish Center for Artificial Intelligence (FCAI), Research Council of Finland flagship program.
We thank the Research Council of Finland for funding the projects ADEREHA (grant no. 353198), BERMUDA (362407), PROFI7 (352788) and MARL (357301).
We also appreciate CSC - IT Center for Science, Finland, for granting access to supercomputers Mahti and Puhti, as well as LUMI, owned by the European High Performance Computing Joint Undertaking (EuroHPC JU) and hosted by CSC Finland in collaboration with the LUMI consortium.
Furthermore, we acknowledge the computational resources provided by the Aalto Science-IT project through the Triton cluster.
\end{acks}

\bibliographystyle{ACM-Reference-Format}
\bibliography{acmart}


\end{document}